\documentclass[aos,citesort,dvips]{arximspdf}
\usepackage{graphics}
\doi{10.1214/09-AOS760}
\volume{38}
\issue{3}
\pubyear{2010}
\firstpage{1665}
\lastpage{1685}

\makeatletter

\newtheorem{prop}{Proposition}[section]

\newproclaim{defn}[prop]{Definition}

\newtheorem{theorem}[prop]{Theorem}

\newproclaim{ex}[prop]{Example}

\newtheorem{cor}[prop]{Corollary}
\newtheorem{lemma}[prop]{Lemma}

\newproclaim{rmk}{Remark}

\newcommand{\indep}{\perp\hspace*{-6.2pt}\perp}

\makeatother

\begin{document}
\begin{frontmatter}

\title{Trek separation for Gaussian graphical models}
\runtitle{Trek separation}

\begin{aug}
\author[A]{\fnms{Seth} \snm{Sullivant}\corref{}\thanksref{t1}\ead[label=e1]{smsulli2@ncsu.edu}},
\author[B]{\fnms{Kelli} \snm{Talaska}\thanksref{t2}\ead[label=e2]{kellicar@umich.edu}} and
\author[C]{\fnms{Jan} \snm{Draisma}\thanksref{t3}\ead[label=e3]{j.draisma@tue.nl}}
\runauthor{S. Sullivant, K. Talaska and J. Draisma}
\affiliation{North Carolina State University, University of Michigan
and Technische~Universiteit Eindhoven}
\address[A]{S. Sullivant\\
Department of Mathematics\\
North Carolina State University\\
Raleigh, North Carolina 27695\\
USA\\
\printead{e1}}
\address[B]{K. Talaska\\
Department of Mathematics\\
University of Michigan\\
Ann Arbor, Michigan 48109-1043 \\
USA\\
\printead{e2}}
\address[C]{J. Draisma\\
Department of Mathematics\\
\quad and Computer Science \\
TU Eindhoven\\
5600 MB Eindhoven\\
The Netherlands\\
and\\
Centrum voor Wiskunde\\
\quad and Informatica\\
Amsterdam\\
The Netherlands\\
\printead{e3}}
\end{aug}

\thankstext{t1}{Supported by NSF Grant DMS-08-40795.}
\thankstext{t2}{Supported by NSF Grants DMS-05-02170 and DMS-05-55880.}
\thankstext{t3}{Supported by DIAMANT, an NWO mathematics cluster.}

% HISTORY:
\received{\smonth{12} \syear{2008}}
\revised{\smonth{9} \syear{2009}}

% ABSTRACT
%
\begin{abstract}
Gaussian graphical models are semi-algebraic subsets of the cone of
positive definite covariance matrices. Submatrices with low rank
correspond to generalizations of conditional independence constraints
on collections of random variables. We give a precise graph-theoretic
characterization of when submatrices of the covariance matrix have
small rank for a general class of mixed graphs that includes directed
acyclic and undirected graphs as special cases. Our new trek
separation criterion generalizes the familiar $d$-separation criterion.
Proofs are based on the trek rule, the resulting matrix factorizations
and classical theorems of algebraic combinatorics on the expansions of
determinants of path polynomials.
\end{abstract}

% KEYWORDS
%

\pdfkeywords{62H99, 62J05, 05A15, Graphical model, Bayesian network, Gessel--Viennot--Lindstrom
lemma, trek rule, linear regression, conditional independence}

\end{frontmatter}

%s1 ###
\section{Introduction}\label{sec:intro}

Given a graph $G$, a graphical model is a family of probability
distributions that satisfy some conditional independence constraints
which are determined by separation criteria in terms of the graph. In
the case of normal random variables, conditional independence
constraints correspond to low rank submatrices of the covariance matrix
$\Sigma$ of a special type. Thus for Gaussian graphical models, the
graphical separation criteria correspond to special submatrices of the
covariance matrix having low rank.

Consider first the case where $G$ is a directed acyclic graph. In this
case, a conditional independence statement $X_A \indep X_B | X_C$ holds
for every distribution consistent with the graphical model if and only
if $C$ $d$-separates $A$ from $B$ in $G$. For normal random variables the
conditional independence constraint $X_A \indep X_B | X_C$ is equivalent
to the condition $\operatorname{rank} \Sigma_{A \cup C, B \cup C} = \#C$
where $\Sigma_{A \cup C, B \cup C}$ is the submatrix of the covariance
matrix $\Sigma$ with row indices $A \cup C$ and column indices $B \cup
C$. However, the drop of rank of a general submatrix $\Sigma_{A,B}$
does not necessarily correspond to a conditional independence statement
that is valid for the graph, and will not, in general, come from a
$d$-separation criterion. Our main result for directed graphical models
is a new separation criterion ($t$-separation) which gives a complete
characterization of when submatrices of the covariance matrix will drop
rank and what the generic lower rank of that matrix will be.

One of the main reasons for searching for necessary and sufficient
conditions for matrices to drop rank comes from the search for a
unified perspective on rank conditions implied by the $d$-separation
criterion and the tetrad representation theorem \cite{Spirtes2000},
which characterizes $2\times2$ vanishing determinants in directed
acyclic graphs. The $t$-separation criterion unifies both of these
results under a simple and more general umbrella.

A second reason for introducing $t$-separation is that it provides a new
set of tools for performing constraint-based inference in Gaussian
graphical models. This approach was pioneered by the TETRAD program
\cite{Scheines1998} where vanishing tetrad constraints are used to
infer the structure of hidden variable graphical models. The
mathematical underpinning of the TETRAD program is the above-mentioned
tetrad representation theorem \cite{Spirtes2000}. In fact, the impetus
for this project was a desire to develop a better understanding of the
tetrad representation theorem. The original proof of the tetrad
representation theorem is lengthy and complicated, and some
simplifications appear in subsequent work \cite
{Shafer1993,Sullivant2008}. Our result has the advantage of being considerably
broader, while our proof is more elementary. The notion that algebraic
determinantal constraints could be useful for inferring graphical
structures is further supported by recent results on the distribution
of the evaluation of determinants of Wishart matrices \cite{Drton2008}
which would be an essential tool for developing Wald-type tests in this setting.

Section \ref{sec:separation} gives the setup of Gaussian graphical
models and states the main results on $t$-separation. To describe the
main result we need to recall the notion of \textit{treks} which are
special paths in the graph $G$. These are the main objects used in the
\textit{trek rule}, a combinatorial parametrization of covariance
matrices that belong to the Gaussian graphical model. We make a special
distinction between general treks and simple treks and introduce two
trek rules. These results are probably well known to experts but are
difficult to find in the literature. Then we make precise the
$t$-separation criterion and state our main results about it. This
section is divided into subsections: stating our results first for
directed graphical models, then undirected graphical models and finally
the more general mixed graphs. The purpose for this division is
twofold: it extracts the two most common classes of graphical models
and it mirrors the structure of the proof of the main results.

Section \ref{sec:proof} is concerned with the proofs of the main
results. The main idea is to exploit the trek rule which expresses
covariances as polynomials in terms of treks in the graph $G$. The
expansion of determinants of matrices of path polynomials is a
classical problem in algebraic combinatorics covered by the
Gessel--Viennot--Lindstr\"{o}m lemma, which we exploit in our proof. The
final tool is Menger's theorem on flows in graphs.

%%%%%%%%%%%%%%%%
%%%%%%%%%%%%%%%%
%%%%%%%%%%%%%%%%

%s2 ###
\section{Treks and $t$-separation}\label{sec:separation}

This section provides background on and definitions of treks as well as
the statements of our main results on $t$-separation for Gaussian
graphical models. We describe necessary and sufficient conditions for
directed and undirected graphs first, and then address the general case
of mixed graphs. The proofs in Section \ref{sec:proof} also follow the
same basic format.

%s2.1 ###
\subsection{Directed graphs}\label{sub:directed}

Let $G$ be a directed acyclic graph with vertex set $V(G) = [m]:= \{
1,2, \ldots, m\}$. We assume $G$ is \textit{topologically ordered}, that
is, we have $i < j$ whenever $i \to j \in E(G)$. A \textit{parent} of a
vertex $j$ is a node $i \in V(G)$ such that $i \to j$ is an edge in
$G$. The set of all parents of a vertex $j$ is denoted $\operatorname{pa}(j)$.
Given such a directed acyclic graph, one introduces a family of normal
random variables that are related to each other by recursive regressions.

To each node $i$ in the graph, we introduce a random variable $X_i$ and
a random variable $\varepsilon_i$. The $\varepsilon_i$ are independent normal
random variables $\varepsilon_i \sim\mathcal{N}( 0, \phi_{i})$ with
$\phi
_i > 0$. We assume that all our random variables have mean zero for
simplicity. The recursive regression property of the DAG gives an
expression for each $X_j$ in terms of $\varepsilon_j$, those $X_i$ with
$i<j$ and some regression parameters $\lambda_{ij}$ assigned to the
edges $i \to j$ in the graph
\[
X_j = \sum_{i \in\operatorname{pa}(j)} \lambda_{ij} X_i + \varepsilon_j.
\]

From this recursive sequence of regressions, one can solve for the
covariance matrix $\Sigma$ of the jointly normal random vector $X$.
This covariance matrix is given by a simple matrix factorization in
terms of the regression parameters and the variance parameters $\phi
_{i}$. Let $\Phi$ be the diagonal matrix $\Phi= \operatorname{diag}(
\phi_{1},
\ldots, \phi_{m})$. Let $L$ be the $m \times m$ upper triangular matrix
with $L_{ij} = \lambda_{ij}$ if $i \to j$ is an edge in $G$, and
$L_{ij} = 0$ otherwise. Set $\Lambda= I - L$ where $I$ is the $m
\times m $ identity matrix.
\begin{prop}[(\cite{Richardson2002}, Section 8)]\label{prop:dirfactor}
The variance--covariance matrix of the normal random variable $X =
\mathcal{N}(0, \Sigma)$ is given by the matrix factorization
\[
\Sigma = \Lambda^{-\top} \Phi\Lambda^{-1}.
\]
\end{prop}

Given two subsets $A, B \subset[m]$, we let $\Sigma_{A,B} = (\sigma
_{ab})_{a \in A, b \in B}$ be the submatrix of covariances with row
index set $A$ and column index set $B$. If $A = B = [m]$, we abbreviate
and say that $\Sigma_{[m], [m]} = \Sigma$. Conditional independence
statements for normal random variables can be detected by investigating
the determinants of submatrices of the covariance matrix \cite{Sullivant2008}.
\begin{prop}\label{prop:condindrank}
Let $X \sim\mathcal{N}(\mu, \Sigma)$ be a normal random variable, and
let $A$, $B$, and $C$ be disjoint subsets of $[m]$. Then the
conditional independence statement $X_A \indep X_B | X_C$ holds for $X$,
if and only if $\Sigma_{A \cup C, B \cup C}$ has rank $ C$.
\end{prop}

Often in the statistical literature, the conditional independence
conditions of a normal random variable are specified by saying that
partial correlations are equal to zero. Proposition \ref
{prop:condindrank} is just an algebraic reformulation of that standard
characterization.

A classic result of the graphical models literature is the
characterization of precisely which conditional independence statements
hold for all densities that belong to the graphical model. This
characterization is determined by the $d$-separation criterion.
\begin{defn}
Let $A$, $B$ and $C$ be disjoint subsets of $[m]$. The set $C$ \textit
{directed separates} or \textit{$d$-separates} $A$ and $B$ if
every path (not necessarily directed) in $G$ connecting a vertex $i \in
A$ to a vertex $j\in B$ contains a vertex $k$ that is either:
\begin{enumerate}
\item a noncollider that belongs to $C$ or
\item a collider that does not belong to $C$ and has no descendants
that belong to $C$,
\end{enumerate}
where $k$ is a \textit{collider} if there exist two edges $a\to k$ and
$b\to k$ on the path and a \textit{noncollider} otherwise.
\end{defn}
\begin{theorem}[(Conditional independence for directed graphical models
\cite{Lauritzen1996})]\label{thm:condgraph}
A set $C$ $d$-separates $A$ and $B$ in $G$ if and only if the conditional
independence statement $X_A \indep X_B | X_C$ holds for every
distribution in the graphical model associated to $G$.
\end{theorem}

Combining Proposition \ref{prop:condindrank} and Theorem \ref
{thm:condgraph} gives a characterization of when all the $(\#C + 1)
\times(\#C + 1)$ minors of a submatrix $\Sigma_{A \cup C, B \cup C}$
must vanish. However, not every vanishing subdeterminant of a
covariance matrix in a Gaussian graphical model comes from a
$d$-separation criterion, as the following example illustrates.
\begin{ex}[(Choke point)]\label{eg:choke}
Consider the graph in Figure \ref{fig:choke} with five vertices and
five edges.
%
%f1 ###
%
\begin{figure}[b]

\includegraphics{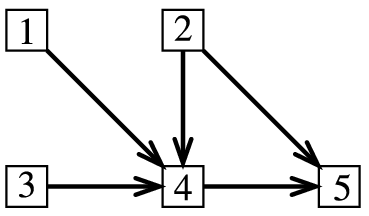}

\caption{}\label{fig:choke}
\end{figure}
In this graph, the determinant $|\Sigma_{13, 45}| = 0$ for any choice
of model parameters. However, this vanishing rank condition does not
follow from any single $d$-separation criterion/conditional independence
statement that is implied by the graph.
\end{ex}

Our main result is an explanation of where these extra vanishing
determinants come from, for Gaussian directed graphical models. Before
we give the precise explanation in terms of treks, we want to first
explain how they enter the story.
\begin{defn}
A \textit{trek} in $G$ from $i$ to $j$ is an ordered pair of directed
paths $(P_1, P_2)$ where $P_1$ has sink $i$, $P_2$ has sink $j$, and
both $P_1$ and $P_2$ have the same source $k$. The common source $k$ is
called the \textit{top} of the trek, denoted $\operatorname
{top}(P_1,P_2)$. Note
that one or both of $P_1$ and $P_2$ may consist of a single vertex,
that is, a path with no edges. A trek $(P_1, P_2)$ is \textit{simple} if
the only common vertex among $P_1$ and $P_2$ is the common source
$\operatorname{top}(P_1, P_2)$. We let $\mathcal{T}(i,j)$ and $\mathcal{S}(i,j)$
denote the sets of all treks and all simple treks from $i$ to $j$, respectively.
\end{defn}

Expanding the matrix product for $\Sigma$ in Proposition \ref
{prop:dirfactor} gives the following \textit{trek rule} for the
covariance $\sigma_{ij}$:
%
%e1 ###
%
\begin{equation}\label{eq:trekrule}
\sigma_{ij} = \sum_{(P_1, P_2) \in\mathcal{T}(i,j)}
\phi_{\mathrm{top}(P_1, P_2)} \lambda^{P_1} \lambda^{P_2},
\end{equation}
where for each path $P$, $\lambda^P$ is the \textit{path monomial} of $P$
defined by
\[
\lambda^P := \prod_{k \to l \in P} \lambda_{kl}.
\]

There is another rule for parameterizing the covariance matrices which
involves sums over only the set $\mathcal{S}(i,j)$ of simple treks. To
describe this, we introduce an alternate parameter $a_i$ associated
to each node $i$ in the graph and defined by the rule
\[
a_i = \sigma_{ii} = \sum_{(P_1, P_2) \in\mathcal{T}(i,i) } \phi
_{\mathrm{top}(P_1, P_2)} \lambda^{P_1} \lambda^{P_2}.
\]
With the definition of the alternate parameter $a_i$, this leads to the
parametrization, called the \textit{simple trek rule},
%
%e2 ###
%
\begin{equation} \label{eq:simptrekrule}
\sigma_{ij} =
\sum_{(P_1, P_2) \in\mathcal{S}(i,j)} a_{\mathrm{top}(P_1, P_2)}
\lambda
^{P_1} \lambda^{P_2}.
\end{equation}
The simple trek rule is also known as Wright's method of path analysis
\cite{Wright1934}.
While we will depend most heavily on the trek rule in this paper, the
simple trek rule also has its uses. In particular, the simple trek rule
played an important role in the study of Gaussian tree models in \cite
{Sullivant2008}.

The fact that treks arise in the expressions for $\sigma_{ij}$ suggests
that any combinatorial rule for the vanishing of a determinant $\Sigma
_{A,B}$ should depend on treks in some way. This leads us to introduce
the following separation criterion that involves treks.
\begin{defn}\label{def:t-sep}
Let $A$, $B$, $C_A$, and $C_B$ be four subsets of $V(G)$ which need not
be disjoint. We say that the pair $(C_A, C_B)$ \textit{trek separates}
(or \textit{$t$-separates}) $A$ from $B$ if for every trek $(P_1,P_2)$ from
a vertex in $A$ to a vertex in $B$, either $P_1$ contains a vertex in
$C_A$ or $P_2$ contains a vertex in $C_B$.
\end{defn}
\begin{rmk*} The following facts follow immediately from
Definition \ref{def:t-sep}:
\begin{enumerate}
\item Since a trek may consist of a single vertex $v$, or more
precisely a pair of paths with zero edges, we must have $A\cap B\subset
C_A\cup C_B$ whenever $(C_A,C_B)$ $t$-separates $A$ from $B$.
\item The pair $(C_A,C_B)$ $t$-separates $A$ from $B$ if and only if the
pair $(C_B,C_A)$ \mbox{$t$-separates} $B$ from $A$.
\item Each of the pairs $(A,\varnothing)$ and $(\varnothing,B)$ always
$t$-separate $A$ from $B$, so we can always find a $t$-separating set of
size $\min(\#A,\#B)$. Our results in this paper will show that
$t$-separation gives nontrivial restrictions on the covariance matrix
when $\#C_A+\#C_B < \min(\#A,\#B)$.
\end{enumerate}
\end{rmk*}

The combinatorial notion of $t$-separation allows us to give a complete
characterization of when submatrices of the covariance matrix can drop
rank. This is the main result for Gaussian directed graphical models;
it will be proved in Section~\ref{sub:pf-directed}.
\begin{theorem}[(Trek separation for directed graphical models)]\label{thm:main}
The submatrix $\Sigma_{A,B}$ has rank less than or equal to $r$ for all
covariance matrices consistent with the graph $G$ if and only if there
exist subsets $C_A,C_B \subset V(G)$ with $\#C_A + \#C_B \leq r$ such
that $(C_A,C_B)$ $t$-separates $A$ from $B$. Consequently,
\[
\operatorname{rk } (\Sigma_{A,B} )\leq\min\{\#C_A+\#C_B\dvtx(C_A,C_B)
\mbox
{ $t$-separates $A$ from $B$}\}
\]
and equality holds for generic covariance matrices consistent with $G$.
\end{theorem}

Here and throughout the paper, the term \textit{generic} means that the
condition holds on a dense open subset of the parameter space. Since
rank conditions are algebraic, this means that the set where the
inequality is strict is an algebraic subset of parameter space with
positive codimension (see \cite{CLO} for background on this algebraic
terminology).
\begin{ex}[(Choke point, continued)] \label{eg:choke2}
Returning to the graph from Example \ref{eg:choke}, we see that
$(\varnothing,\{4\})$ $t$-separates $\{1,3\}$ from $\{4,5\}$ which implies
that the submatrix $\Sigma_{13, 45}$ has rank at most one for every
matrix that belongs to the model. Thus $t$-separation explains this extra
vanishing minor that $d$-separation misses.
\end{ex}

Readers familiar with the tetrad representation theorem will recognize
that $\{4\}$ is a choke point between $\{1,3\}$ and $\{4,5\}$ in $G$.
In particular, Theorem \ref{thm:main} includes the tetrad
representation theorem as a special case.
\begin{cor}[(Tetrad representation Theorem \cite{Spirtes2000})]
The tetrad $\sigma_{ik}\sigma_{jl} - \sigma_{il} \sigma_{jk}$ is zero
for all covariance matrices consistent with the graph $G$ if and only
if there is a node $c$ in the graph such that either $(\{c\},
\varnothing
)$ or $(\varnothing, \{c\})$ $t$-separates $\{i,j\}$ from $\{k,l\}$.
\end{cor}

Since conditional independence in a directed graphical model
corresponds to the vanishing of subdeterminants of the covariance
matrix, the $t$-separation criterion can be used to characterize these
conditional independence statements, as well.
\begin{theorem}\label{thm:condindtrek}
The conditional independence statement $X_A \indep X_B | X_C$ holds for
the graph $G$ if and only if there is a partition $C_A \cup C_B = C$ of
$C$ such that $(C_A,C_B)$ $t$-separates $A \cup C$ from $B \cup C$ in $G$.
\end{theorem}
\begin{pf}
The conditional independence statement holds for the graph $G$ if and
only if the submatrix of the covariance matrix $\Sigma_{A \cup C, B
\cup C}$ has rank $\# C$. By trek separation for directed graphical
models, this holds if and only if there exists a pair of sets $D_A$ and
$D_B$, with $\# D_A + \#D_B = \# C$ such that $(D_A,D_B)$ $t$-separates
$A \cup C$ from $B \cup C$. Among the treks from $A \cup C$ to $B \cup
C$ are the lone vertices $c \in C$. Hence $C \subseteq D_A \cup D_B$.
Since $\#D_A + \#D_B = \#C$, we must have $D_A \cup D_B = C$ and these
two sets form a partition of $C$.
\end{pf}

Theorem \ref{thm:condindtrek} immediately implies that $d$-separation is
a special case of \mbox{$t$-separation}. Yanming Di \cite{Di2009} found a direct
combinatorial proof of this fact after we made a preliminary version of
this paper available.
\begin{cor}
A set $C$ $d$-separates $A$ and $B$ in $G$ if and only if there is a
partition $C = C_A \cup C_B$ such that $(C_A, C_B)$ $t$-separates $A \cup
C$ from $B \cup C$.
\end{cor}

While $t$-separation includes $d$-separation, and the vanishing minors of
conditional independence, as a special case, it also seems to capture
some new vanishing minor conditions that do not follow from
$d$-separation. The most interesting cases of this seem to occur when
$C_A \cap C_B \neq\varnothing$.
\begin{ex}[(Spiders)]\label{eg:spiders}
Consider the graph in Figure \ref{fig:spider} which we call a \textit{spider}.

%f2 ###
%
\begin{figure}[b]

\includegraphics{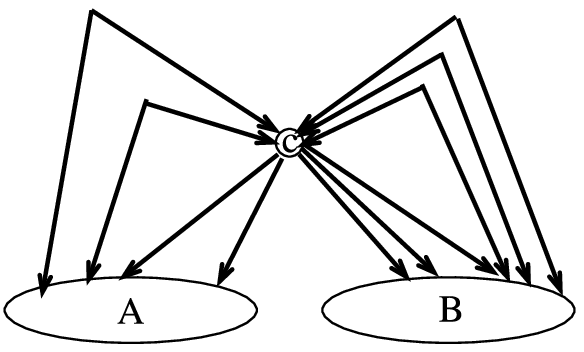}

\caption{}\label{fig:spider}
\end{figure}

Clearly, we have that $(\{c\},\{c\})$ $t$-separates $A$ from $B$, so that
the submatrix $\Sigma_{A,B}$ has rank at most $2$. Although this rank
condition must be implied by CI rank constraints on $\Sigma$ and the
fact that $\Sigma$ is positive definite, it does not appear to be
easily derivable from these constraints.
\end{ex}

%%%%%%%%%%%%%%%%%%%%%%%%%%%%%%%%%
%%%%%%%%%%%%%%%%%%%%%%%%%%%%%%%%%
%%%%%%%%%%%%%%%%%%%%%%%%%%%%%%%%%
%%%%%%%%%%%%%%%%%%%%%%%%%%%%%%%%%

%s2.2 ###
\subsection{Undirected graphs}\label{sub:undirected}

For Gaussian undirected graphical models, the allowable covariance
matrices are specified by placing restrictions on the entries of the
concentration matrix. In particular, let $G$ be an undirected graph,
with edge set $E$. We consider all covariance matrices $\Sigma$ such
that $(\Sigma^{-1})_{ij} = 0$ for all $i -j \notin E(G)$.

As in the case of directed acyclic graphs, it is known that conditional
independence constraints characterize the possible probability
distributions for positive densities \cite{Lauritzen1996}. Indeed, in
the Gaussian case, the pairwise constraints $X_i \indep X_j | X_{[m]
\setminus\{i,j\}}$ for $i -j \notin E(G)$ characterize the
distributions that belong to the model. As in the case of directed
graphical models, general conditional independence constraints $X_A
\indep X_B | X_C$ are characterized by a separation criterion.

If $A$, $B$ and $C$ are three subsets of vertices of an undirected
graph $G$, not necessarily disjoint, we say that $C$ \textit{separates}
$A$ and $B$ if every path from a vertex in $A$ to a vertex in $B$
contains some vertex of $C$.
\begin{theorem}[(Conditional independence for undirected graphical
models~\cite{Lauritzen1996})] For disjoint subsets $A, B, $ and $C
\subseteq[m]$ the conditional independence statement $X_A \indep X_B |
X_C$ holds for the graph $G$ if and only if $C$ separates $A$ and~$B$.
\end{theorem}

Since conditional independence for normal random variables corresponds
to the vanishing of the minors of submatrices of the form $\Sigma_{A
\cup C , B \cup C}$ it is natural to ask what conditions determine the
vanishing of an arbitrary minor $\Sigma_{A,B}$. We will show that the
path separation criterion also characterizes the vanishing of arbitrary
minors for the undirected graphical model.
\begin{theorem}\label{thm:undirected}
The submatrix $\Sigma_{A,B}$ has rank less than or equal to
$r$ for all covariance matrices consistent
with the graph $G$ if and only if there is a set $C \subseteq V(G)$
with $\#C \leq r$ such that $C$ separates $A$ and $B$. Consequently,
\[
\operatorname{rk} (\Sigma_{A,B} )\leq\min\{\#C\dvtx C\mbox{ separates
$A$ and $B$}\}
\]
and equality holds for generic covariance matrices consistent with $G$.
\end{theorem}

Note that the sets $A,B $ and $C$ need not be disjoint in Theorem \ref
{thm:undirected}. We will provide a proof of Theorem \ref
{thm:undirected} in Section \ref{sub:pf-undirected}, using the
combinatorial expansions of determinants. Unlike in the case of
directed acyclic graphs, we do not find any new constraints that were
not trivially implied by conditional independence.

%In an undirected graphical model, every vanishing determinant of a
%submatrix $\Sigma_{A,B}$ is directly implied by a conditional
%independence statement that holds in the graph.

%%%%%%%%%%%%%%%%%%%%%%%%%%%%%%%%%
%%%%%%%%%%%%%%%%%%%%%%%%%%%%%%%%%
%%%%%%%%%%%%%%%%%%%%%%%%%%%%%%%%%
%%%%%%%%%%%%%%%%%%%%%%%%%%%%%%%%%

%s2.3 ###
\subsection{Mixed graphs}\label{sub:mixed}

In this section, we describe our results for general classes of mixed
graphs, that is, graphs that can involve directed edges $i \to j$,
undirected edges $i -j$ and bidirected edges $i \leftrightarrow j$.
We assume that in our mixed graphs there is a partition of the vertices
of the graph $U \cup W = V(G)$, such that all undirected edges have
their vertices in $U$, all bidirected edges have their vertices in $W$
and any directed edge with a vertex in $U$ and a vertex in $W$ must be
of the form $u \to w$ where $u \in U$ and $w \in W$. With all of these
assumptions on our mixed graph, we can order the vertices in such a way
that all vertices in $U$ come before the vertices in $W$, and whenever
$i \to j$ is a directed edge, we have $i < j$. We assume that the
subgraph on directed edges in acyclic. Note that we allow a pair of
vertices to be connected by both a directed edge $i \to j$ and a
bidirected edge $i \leftrightarrow j$ or undirected edge $i -j$.
With this setup, both ancestral graphs \cite{Richardson2002} and chain
graphs \cite{AMP2001} occur as special cases.

Now we introduce three matrices which are determined by the three
different types of edges in the graph. We first let $\Lambda$ be the
matrix with rows and columns indexed by $V(G)$ which is defined by
$\Lambda_{ii}=1$, $\Lambda_{ij} = - \lambda_{ij}$ if $i \to j \in E(G)$
and $\Lambda_{ij} = 0$ otherwise. Each $\lambda_{ij}$ is a real
parameter associated to a directed edge in $G$, though they no longer
necessarily have the interpretation of regression coefficients. Next,
we let $K$ be a symmetric positive definite matrix, with rows and
columns indexed by $U$, such that $K_{ij} = 0$ if $i -j \notin
E(G)$. Each entry $K_{ij}$ with $i \neq j$ is a parameter associated
to an undirected edge in $G$. Finally, we let $\Phi= (\phi_{ij})$ be
a symmetric positive definite matrix, with rows and columns indexed by
$W$, such that $\phi_{ij} = 0$ if $i \leftrightarrow j \notin E(G)$.
Each $\phi_{ij}$ with $i \neq j$ is a parameter associated to a
bidirected edge in $G$.

From the three matrices $\Lambda$, $K$ and $\Phi$, defined as above, we
obtain the following covariance matrix of our mixed graphical model:
\[
\Sigma= \Lambda^{-\top}
\pmatrix{
K^{-1} & 0 \cr
0 & \Phi}
\Lambda^{-1}.
\]
Note that this representation parametrizes the Gaussian ancestral graph
model in the case where $G$ is an ancestral graph \cite
{Richardson2002}, and chain graph models under the alternative Markov
property \cite{AMP2001}, when $G$ is a chain graph.

We use a path expansion in Section \ref{sub:pf-mixed} to express this
factorization as a power series of sums of paths, analogous to the
polynomial expressions in terms of treks that appeared in the purely
directed case in Section \ref{sub:directed}. In the precise formulation
given in Section \ref{sub:pf-mixed}, we will need the following
generalized notion of a trek.

A \textit{trek} between vertices $i$ and $j$ in a mixed graph $G$ is a
triple $(P_L, P_M, P_R)$ of paths where:
\begin{enumerate}
\item$P_L$ is a directed path of directed edges with sink $i$;
\item$P_R$ is a directed path of directed edges with sink $j$;
\item$P_M$ is either:
\begin{itemize}
\item a path consisting of zero or more undirected edges connecting the
source of $P_L$ to the source of $P_R$, or
\item a single bidirected edge connecting the source of $P_L$ to the
source of $P_R$.
\end{itemize}
\end{enumerate}
A trek $(P_L, P_M, P_R)$ is called \textit{simple} if each of $P_L$,
$P_M$ and $P_R$ is self-avoiding, and the only vertices which appear in
more than one of the segments $P_L$, $P_M$, and~$P_R$ are the sources
of $P_L$ and $P_R$.

The set of all treks between $i$ and $j$ is denoted by $\mathcal
{T}(i,j)$ and the set of all simple treks is $\mathcal{S}(i,j)$. Note
that $\mathcal{T}(i,j)$ might be infinite because we allow the path
$P_M$ to have cycles. On the other hand, $\mathcal{S}(i,j)$ is always finite.
\begin{defn}
A triple of sets of vertices $(C_L, C_M, C_R)$ \textit{$t$-separates} $A$
from $B$ in the mixed graph $G$ if for every simple trek $(P_L, P_M,
P_R)$ with the sink of $P_L$ in $A$ and the sink of $P_R$ in $B$, we
have that $P_L$ contains a vertex in $C_L$, $P_R$ contains a vertex in
$C_R$ or $P_M$ is an undirected path that contains a vertex in $C_M$.
\end{defn}

Note that the mixed graph definition of $t$-separation reduces to the
directed acyclic graph version of $t$-separation when $G$ is a DAG and
reduces to ordinary graph separation when $G$ is an undirected graph.
\begin{theorem}[($t$-separation for mixed graphs)]\label{thm:mixed}
The matrix $\Sigma_{A,B}$ has rank at most $r$ for all covariance
matrices consistent with the mixed graph $G$ if and only if there exist
three subsets $C_L, C_M, C_R$ with $\#C_L + \#C_M + \#C_R \leq r$ such
that $(C_L, C_M, C_R)$ $t$-separates $A$ from $B$. Consequently,
\[
\operatorname{rk} (\Sigma_{A,B} )\leq\min\{\#C_L+\#C_M+\#C_R\dvtx
(C_L,C_M,C_R) \mbox{ $t$-separates $A$ from $B$}\}
\]
and equality holds for generic covariance matrices consistent with $G$.
\end{theorem}

Since conditional independence statements for Gaussian graphical models
correspond to special low rank submatrices of the covariance matrix,
Theorem \ref{thm:mixed} also gives a characterization of when
conditional independence statements for these mixed graph models hold.
\begin{cor}
The conditional independence statement $X_A \indep X_B | X_C$ holds for
the Gaussian graphical model associated to the mixed graph $G$, if and
only if there is a partition $C = C_L \cup C_M \cup C_R$ such that
$(C_L, C_M, C_R)$ $t$-separates $A \cup C$ from $B \cup C$.
\end{cor}
\begin{pf}
The conditional independence statement holds if and only if\break $\Sigma_{A
\cup C, B \cup C}$ has rank $\#C$. By Theorem \ref{thm:mixed} this
happens if and only there exists $(D_L, D_M, D_R)$ with $\#D_L + \#D_M
+ \#D_R \leq\#C$ that $t$-separate $A \cup C$ and $B \cup C$. But since
$C \subseteq D_L \cup D_M \cup D_R$, this occurs if and only if $C =
D_L \cup D_M \cup D_R$ is a partition of $C$.
\end{pf}

It is worth noting, however, that unlike in the case of directed
acyclic graphs and undirected graphs, conditional independence
statements and vanishing minors are not enough to characterize the
covariance matrices that come from the model. See the example in
Section 8.3.1 of \cite{Richardson2002}.

%%%%%%%%%%%%%%%%%%%%%%%%%%%%%%%%%
%%%%%%%%%%%%%%%%%%%%%%%%%%%%%%%%%
%%%%%%%%%%%%%%%%%%%%%%%%%%%%%%%%%
%%%%%%%%%%%%%%%%%%%%%%%%%%%%%%%%%

%s3 ###
\section{Proofs}\label{sec:proof}

In this section, we consider the elements $\lambda_{ij}$, $\phi_{ij}$
and $k_{ij}$ as polynomial variables or indeterminates. When we speak
about $\det\Sigma_{A,B}$ we mean to speak of this polynomial as an
algebraic object without reference to its evaluation at specific values
of $\lambda_{ij}$, $\phi_{ij}$ and $k_{ij}$. Thus the statement that
$\det\Sigma_{A,B}$ is identically equal to zero means that the
determinant is equal to the zero polynomial or power series.

%s3.1 ###
\subsection{\texorpdfstring{Proof of Theorem \protect\ref{thm:main} (directed
graphs)}{Proof of Theorem 2.8 (directed graphs)}}\label{sub:pf-directed}

Let $G$ be a directed acyclic graph with vertex set $V(G)=[m]$. We
assign to each edge $i \to j$ in $G$ the parameter~$\lambda_{ij}$. Let
$L$ be the $m\times m$ matrix given by $L_{ij}=\lambda_{ij}$ if $i\to
j$ is an edge in $G$ and $L_{ij}=0$ otherwise. Set $\Lambda=I-L$, where
$I$ is the $m\times m$ identity matrix. We assign to each vertex $i\in
[m]$ the parameter $\phi_i$, and let $\Phi$ be the diagonal matrix
$\Phi= \operatorname{diag}(\phi_1, \ldots, \phi_m)$.

The entries of the matrix $\Lambda^{-1}$ have a well-known
combinatorial interpretation in terms of the directed acyclic graph $G$.
\begin{prop} \label{prop:classic} For each path $P$ in the directed
acyclic graph $G$, set $\lambda^P = \prod_{k \to l \in P} \lambda
_{kl}$. Then
\[
(\Lambda^{-1})_{ij} = \sum_{P \in\mathcal{P}(i,j)} \lambda^P,
\]
where $\mathcal{P}(i,j)$ is the set of all directed paths from $i$ to $j$.
\end{prop}
\begin{lemma}\label{lem:CB}
Suppose that $A, B \subseteq[m]$ with $\#A = \#B$. Then $\det\Sigma
_{A,B}$ is identically zero if and only if for every set $S \subset
[m]$ with $\#S = \#A = \#B$, either $\det(\Lambda^{-1})_{S,A} = 0$ or
$\det(\Lambda^{-1})_{S,B} = 0$.
\end{lemma}
\begin{pf}
Since $\Sigma= \Lambda^{-\top} \Phi\Lambda^{-1}$, we have
$\Sigma_{A,B} = (\Lambda^{-\top})_{A,[m]} \Phi(\Lambda^{-1})_{[m],B}$.
We can calculate $\det\Sigma_{A,B}$ by applying the Cauchy--Binet
determinant expansion formula twice on this product. In particular, we obtain
\[
\det\Sigma_{A,B} = \sum_{R,S\subseteq[m]} \det(\Lambda^{-\top})_{A,R}
\det\Phi_{R,S} \det(\Lambda^{-1})_{S,B},
\]
where the sum runs over subsets $R$ and $S$ of cardinality $\#A = \#B$.
Since $\Phi$ is a diagonal matrix, $\det\Phi_{R,S} = 0$ unless $R =
S$, in which case we let $\phi_S$ denote $\det\Phi_{S,S} = \prod_{s
\in S} \phi_s$.

Thus we have the following expansion of $\det\Sigma_{A,B}$:
\begin{eqnarray*}
\det\Sigma_{A,B} &=& \sum_{S \subseteq[m]} \det(\Lambda^{-\top
})_{A,S} \det(\Lambda^{-1})_{S,B} \phi_S\\
&=& \sum_{S \subseteq[m]} \det(\Lambda^{-1})_{S,A} \det(\Lambda
^{-1})_{S,B} \phi_S.
\end{eqnarray*}
Since each monomial $\phi_S$ appears in only one term in this
expansion, the result follows.
\end{pf}

To prove the main theorem, we need two classical results from
combinatorics. The first is Lemma \ref{lem:gvl}, the
Gessel--Viennot--Linstr\"{o}m lemma, which gives a combinatorial
expression for expansions of subdeterminants of the matrix $\Lambda
^{-1}$. The second is Theorem \ref{thm:menger}, Menger's theorem, which
describes a relationship between nonintersecting path families and
blocking sets in a graph.
\begin{lemma}[(Gessel--Viennot--Lindstr\"{o}m lemma \cite
{GV1985,Lindstrom1973})]\label{lem:gvl}
Suppose $G$ is a directed acyclic graph with vertex set $[m]$. Let $R$
and $S$ be subsets of $[m]$ with $\#R = \#S = \ell$. Then
\[
\det(\Lambda^{-1})_{R,S} = \sum_{\mathbf{P}\in N(R,S)} (-1)^\mathbf
{P}\lambda
^\mathbf{P},
\]
where $N(R,S)$ is the set of all collections of nonintersecting
systems of $\ell$ directed paths in $G$ from $R$ to $S$, and
$(-1)^\mathbf{P}
$ is the sign of the induced permutation of elements from $R$ to $S$.
In particular, $\det(\Lambda^{-1})_{R,S} = 0$ if and only if every
system of $\ell$ directed paths from $R$ to $S$ has two paths which
share a vertex.
\end{lemma}

Consider a system $\mathbf{T}=\{T_1,\ldots, T_\ell\}$ of $\ell$
treks from
$A$ to $B$, connecting $\ell$ distinct vertices $a_i\in A$ to $\ell$
distinct vertices $b_j\in B$. Let $\operatorname{top}(\mathbf{T})$
denote the multiset
$\{\operatorname{top}(T_1),\ldots, \operatorname{top}(T_\ell)\}$. Note
that $\mathbf{T}$
consists of two systems of directed paths, a path system $\mathbf
{P}_A$ from
$\operatorname{top}(\mathbf{T})$ to $A$ and a path system $\mathbf
{P}_B$ from
$\operatorname{top}(\mathbf{T}
)$ to $B$. We say that $\mathbf{T}$ has a \textit{sided intersection}
if two
paths in $\mathbf{P}_A$ share a vertex or if two paths in $\mathbf
{P}_B$ share a vertex.
\begin{prop}\label{prop:trekcross}
Let $A$ and $B$ be subsets of $[m]$ with $\#A=\#B$. Then
\[
\det\Sigma_{A,B} = 0,
\]
if and only if every system of (simple) treks from $A$ to $B$ has a
sided intersection.
\end{prop}
\begin{pf}
Suppose that $\det\Sigma_{A,B} = 0$, and let $\mathbf{T}$ be a trek system
from $A$ to $B$. If all elements of the multiset $\operatorname
{top}(\mathbf
{T})$ are
distinct, then Lemma \ref{lem:CB} implies that either $\det(\Lambda
^{-1})_{\operatorname{top}(\mathbf{T}),A}=0$ or $\det(\Lambda
^{-1})_{\operatorname{top}(\mathbf{T}
),B}=0$. If $\operatorname{top}(\mathbf{T})$ has repeated elements,
then these
determinants are zero, since there are repeated rows. Then Lemma \ref
{lem:gvl} implies that there is an intersection in the path system from
$\operatorname{top}(\mathbf{T})$ to $A$ or in the path system from
$\operatorname{top}(\mathbf{T})$
to $B$ which means that $\mathbf{T}$ has a sided intersection.

Conversely, suppose that every trek system $\mathbf{T}$ from $A$ to
$B$ has a
sided intersection, and let $R\subseteq[m]$ with $\#R=\#A=\#B$. If
$R=\operatorname{top}(\mathbf{T})$ for some trek system $\mathbf{T}$
from $A$
to $B$, then
either the path system from $\operatorname{top}(\mathbf{T})$ to $A$ or
the path system
from $\operatorname{top}(\mathbf{T})$ to $B$ has an intersection. If
$R$ is not the
set of top elements for some trek system $\mathbf{T}$, then there is
no path
system connecting $R$ to $A$ or there is no path system connecting $R$
to $B$. In both cases, Lemma \ref{lem:gvl} implies that either $\det
(\Lambda^{-1})_{R,A}=0$ or $\det(\Lambda^{-1})_{R,B}=0$. Lemma \ref
{lem:CB} then implies that $\det\Sigma_{A,B} = 0$.

We note that it is sufficient to check the systems of simple treks.
Given a trek $T$ from $i$ to $j$, let $\operatorname{LE}(T)$ denote
the unique simple
trek from $i$ to $j$ whose edge set is a subset of the edge set of $T$.
Now if each simple trek system $\mathbf{T}$ has a sided intersection, then
every trek system does, namely the intersection coming from
$\operatorname{LE}(\mathbf{T})$.
\end{pf}

We define a new DAG associated to $G$, denoted $\widetilde{G}$, which
has $2m$ vertices $\{1, 2, \ldots, m\} \cup\{1', 2', \ldots, m'\}$ and
edges $i \to j$ if $i \to j$ is an edge in $G$, $j' \to i'$ if $i \to
j$ is an edge in $G$ and $i' \to i$ for each $i \in[m]$.
\begin{prop}\label{prop:trektopath}
Treks in $G$ from $i$ to $j$ are in bijective correspondence with
directed paths from $i'$ to $j$ in $\widetilde{G}$. Simple treks in $G$
from $i$ to $j$ are in bijective correspondence with directed paths
from $i'$ to $j$ in $\widetilde{G}$ that use at most one edge from any
pair $a\to b$ and $b' \to c'$ where $a,b,c\in[m]$.
\end{prop}
\begin{pf}
Every trek is the union of two paths with a common top. The part of the
trek from the top to $i$ corresponds to the subpath with only vertices
in $\{1', \ldots, m'\}$, and the part of the trek from the top to $j$
corresponds to the subpath with only vertices in $\{1, \ldots, m\}$.
The unique edge of the form $k' \to k$ corresponds to the top of the
trek. Excluding pairs $a\to b$ and $b' \to c'$ implies that a trek
never visits the same vertex $b$ twice.
\end{pf}

Menger's theorem (or, more generally, the Max-Flow--Min-Cut theorem) now
allows us to turn our sided crossing result on $G$ into a blocking
characterization on~$\widetilde{G}$.
\begin{theorem}[(Vertex version of Menger's theorem)]\label{thm:menger}
The cardinality of the largest set of vertex disjoint directed paths
between two nonadjacent vertices $u$ and $v$ in a directed graph is
equal to the cardinality of the smallest \textit{blocking set} where a
blocking set is a set of vertices whose removal from the graph ensures
that there is no directed path from $u$ from $v$.
\end{theorem}
\begin{pf*}{Proof of Theorem \protect\ref{thm:main}}
We first focus on the case where $\det\Sigma_{A,B} = 0$ so that the
rank is at most $k-1$ where $k=\#A=\#B$. According to Proposition \ref
{prop:trekcross}, every system of $k$ treks from $A$ to $B$ must have a
sided intersection. That is, the number of vertex disjoint paths from
$A'$ to $B$ is at most $k-1$ in the graph~$\widetilde{G}$. We add two
new vertices to $\widetilde{G}$, one vertex $u$ that points to each
vertex in $A'$ and one vertex $v$ such that each vertex in $B$ points
to $v$. Thus there are at most $k-1$ vertex disjoint paths from $u$ to
$v$. Applying Menger's theorem, there is a blocking set $W$ in
$\widetilde{G}$ of cardinality $k-1$ or less. Set $C_A=\{i\in[m]\dvtx
i'\in W\}$ and $C_B=\{i\in[m]\dvtx i\in W\}$. Then it is clear that $\#
C_A+\#C_B\leq k-1$, and these two sets $t$-separate $A$ from $B$.

Conversely, suppose there exist sets $C_A$ and $C_B$ with $\#C_A+\#
C_B\leq k-1$ which $t$-separate $A$ from $B$. Then $W=\{i\dvtx i\in C_B\}
\cup
\{i'\dvtx i\in C_A\}$ is a blocking set between $u$ and $v$ as above.
Applying Menger's theorem, since $\#W\leq k-1$, there is no vertex
disjoint system of $k$ paths from $A'$ to $B$. Thus every trek system
from $A$ to $B$ will have a sided intersection, so that $\det\Sigma
_{A,B} = 0$ by Proposition \ref{prop:trekcross}.

From the special case of determinants, we deduce the general result,
because if the smallest blocking set has size $r$, there exists a
collection of $r$ disjoint paths between any subset of $A$ and any
subset of $B$, and this is the largest possible number of paths in such
a collection. This means that all $(r+1) \times(r+1)$ minors of
$\Sigma
_{A,B}$ are zero, but at least one $r \times r$ minor is not zero.
Hence $\Sigma_{A,B}$ has rank $r$ for generic choices of the $\lambda$
and $\phi$ parameters.
\end{pf*}

%%%%%%%%%%%%%%%%%%%%%%%%%%%%%%%%%
%%%%%%%%%%%%%%%%%%%%%%%%%%%%%%%%%
%%%%%%%%%%%%%%%%%%%%%%%%%%%%%%%%%
%%%%%%%%%%%%%%%%%%%%%%%%%%%%%%%%%

%s3.2 ###
\subsection{\texorpdfstring{Proof of Theorem \protect\ref{thm:undirected} (undirected
graphs)}{Proof of Theorem 2.15 (undirected graphs)}}\label{sub:pf-undirected}

To prove Theorem \ref{thm:undirected}, we will introduce Lemma \ref
{lem:gvl-undirected}, a limited analogue of the
Gessel--Viennot--Lindstr\"{o}m lemma for graphs which are not
necessarily acyclic. This version is a direct corollary of Theorem 6.1
in \cite{Fomin2001} which, for the sake of simplicity, we do not state
in full generality.

Let $G$ be a directed graph, not necessarily acyclic. Let $W$ be the
matrix given by $W_{ij}=w_{ij}$ if $i\rightarrow j$ is an edge in $G$
and $W_{ij}=0$ otherwise. By standard notions in algebraic graph
theory, we can expand the matrix $(I - W)^{-1}$ as a formal power
series in terms of the $w_{ij}$. In particular,
\[
(I - W)^{-1}_{ij} = \sum_{P \in\mathcal{P}(i,j)} w^P,
\]
where $\mathcal{P}(i,j)$ is the set of all (possibly infinitely many)
paths from $i$ to $j$ in $G$. This is just Proposition \ref
{prop:classic} in the general case.

Let $A=\{a_1, \ldots, a_\ell\}$ and $B=\{b_1, \ldots, b_\ell\}$ be
subsets of $[m]$ with the same cardinality. The determinant $\det((I-
W)^{-1})_{A,B}$ can be written simply in an expression that involves
cancelation as
%
%e3 ###
%
\begin{equation}\label{eq:bigexpand}
\det\bigl((I- W)^{-1}\bigr)_{A,B} = \sum_{\tau\in S_\ell, P_i \in\mathcal
{P}(a_{i}, b_{\tau(i)})} \operatorname{sign}(\tau) \prod_{i=1}^\ell w^{P_i}.
\end{equation}
Deciding whether this formula is nonzero amounts to showing whether or
not all terms cancel in this formula. This leads to the following
version of the Gessel--Viennot--Lindstr\"{o}m lemma \cite{Fomin2001}.
\begin{lemma}\label{lem:gvl-undirected}
Let $G$ be a directed graph. Let $A=\{a_1, \ldots, a_\ell\}$ and $B=\{
b_1, \ldots, b_\ell\}$ be subsets of $[m]$ with the same cardinality.
Then $(\det(I-W)^{-1})_{A,B}$ is identically zero if and only if every
system of $\ell$ directed paths from $A$ to $B$ has two paths which
share a vertex. Further, if there is a set of $\ell$ paths $P_1,
\ldots
, P_\ell$ from $A$ to $B$ which do not have a common vertex, then
$w^{P_1} \cdots w^{P_\ell}$ appears as a monomial with a nonzero
coefficient in the power series expansion of $\det( (I-W)^{-1})_{A,B}$.
\end{lemma}

For an undirected graph $G$, we associate to each edge $i-j$ in $G$
a parameter~$\psi_{ij}$. Then let $\Psi_{ij}=\psi_{ij}$ if $i-j$ is
an edge in $G$ and $\Psi_{ij}=0$ otherwise. Let $\widehat{G}$ be the
directed graph formed by replacing each undirected edge in $G$ with two
directed edges of weight $\psi_{ij}$, one in each direction.
\begin{cor}\label{cor:ghat-gvl}
For this symmetric matrix $\Psi$, the determinant $\det((I-\Psi
)^{-1})_{A,B}$ is identically\vspace*{1pt} zero if and only if every system of $\ell
= \#A = \#B$ directed paths from $A$ to $B$ in $\widehat{G}$ has two
paths which share a vertex.
\end{cor}
\begin{pf}
Lemma \ref{lem:gvl-undirected} immediately implies that if every system
of directed paths in $\widehat{G}$ has a crossing, then $\det((I-\Psi
)^{-1})_{A,B}$ is identically zero, by specialization.

To show the converse, we need to verify that, for a fixed $A$ and $B$,
each system $\mathbf{P}$ consisting of self-avoiding paths, no two of which
intersect, is the unique system of its weight $\psi^\mathbf{P}$. While
$\widehat{G}$ may have multiple path systems of the same weight $\psi
^\mathbf{P}$, they must all consist of the same undirected edges in
$G$, and
any such system in $\widehat{G}$ can be obtained from any other by
switching the directions of some of the paths. Then, since no two of
the paths intersect, we see that there is only one such system with the
correct orientation of paths, since $A$ and $B$ are fixed.
\end{pf}
\begin{pf*}{Proof of Theorem \protect\ref{thm:undirected}}
We write $\Sigma= K^{-1} =D^{-1}(I-\Psi)^{-1}D^{-1}$
where $D$ is the diagonal matrix of standard deviations, $D =
\operatorname{diag}(\sqrt{\sigma_{11}}, \ldots,\break \sqrt{\sigma_{mm}})$.
We can treat
the entries $\Psi_{ij} = k_{ij}\cdot\sqrt{\sigma_{ii} \sigma
_{jj}}$ as
free parameters. It suffices to prove a vanishing determinant condition
locally near a single point in the parametrization, so we assume that
$\Psi$ is small so that we can use the power series expansion,
$(I-\Psi
)^{-1}=I+\Psi+\Psi^2+\Psi^3+\cdots.$
Applying Cauchy--Binet as before, we obtain
\begin{eqnarray*}
\det\Sigma_{A,B} &=& \sum_{R,S\subseteq[m]} \det(D^{-1})_{A,R} \det
\bigl((I-\Psi)^{-1}\bigr)_{R,S} \det(D^{-1})_{S,B}\\
&=& \det(D^{-1})_{A,A} \det\bigl((I-\Psi)^{-1}\bigr)_{A,B} \det(D^{-1})_{B,B},
\end{eqnarray*}
since $\det(D^{-1})_{A,R}=0$ if $A\neq R$ and $\det(D^{-1})_{S,B}=0$ if
$B\neq S$. Now,\break $\det(D^{-1})_{A,A} \neq0$ and $\det(D^{-1})_{B,B}
\neq0$, and Corollary \ref{cor:ghat-gvl} completes the proof.
%\rightqed
\end{pf*}

%%%%%%%%%%%%%%%%%%%%%%%%%%%%%%%%%
%%%%%%%%%%%%%%%%%%%%%%%%%%%%%%%%%
%%%%%%%%%%%%%%%%%%%%%%%%%%%%%%%%%
%%%%%%%%%%%%%%%%%%%%%%%%%%%%%%%%%

%s3.3 ###
\subsection{\texorpdfstring{Proof of Theorem \protect\ref{thm:mixed} (mixed
graphs)}{Proof of Theorem 2.17 (mixed graphs)}}\label{sub:pf-mixed}

Recall that covariance matrices consistent with a mixed graph $G$ all
have the form
\[
\Sigma= \Lambda^{-\top}
\pmatrix{
K^{-1} & 0 \cr
0 & \Phi
}
\Lambda^{-1}.
\]

Our first step is a standard argument in the graphical models
literature, which allows us to reduce to the case where there are no
bidirected edges in the graph. This can be achieved by subdividing the
bidirected edges; that is, for each bidirected edge $i \leftrightarrow
j$ in the graph, where $i\leq j$, we replace $i\leftrightarrow j$ with
a vertex $v_{i,j}$, directed edges $v_{i,j} \to i$ and $v_{i,j} \to j$.
The graph $\widetilde{G}$ obtained from $G$ by subdividing all of its
bidirected edges is called the \textit{bidirected subdivision} of $G$.
If $G$ has only directed and bidirected edges, then $\widetilde{G}$ is
called the canonical DAG associated to $G$.
\begin{prop}\label{prop:removebi}
Let $A,B \subset V(G)$ be two sets of vertices such that $\#A=\#B$.
\begin{enumerate}
\item The generic rank of $\Sigma_{A,B}$ is the same for matrices
compatible with $G$ or $\widetilde{G}$.
\item There exists a triple $(C_L, C_M, C_R)$ with $\#C_L + \#C_M + \#
C_R = r$ that $t$-separates $A$ from $B$ in $G$ if and only if there is a triple
$(D_L, D_M, D_R)$ with $\#D_L + \#D_M + \#D_R = r$ that $t$-separates $A$
from $B$ in $\widetilde{G}$.
\end{enumerate}
\end{prop}
\begin{pf}
(1) It suffices to prove that the two parametrizations have the same
Zariski closure (see \cite{CLO} for the definition and background).
This will follow by showing that near the identity matrix, the two
parameterizations give the same family of matrices. Locally near the
identity matrix, the matrix expansion for $\Sigma$ can be expanded as a
formal power series in the entries of $K$, $\Phi$ and $\Lambda$. The
expansion for $\sigma_{ij}$ can be expressed as a sum over all treks
$\mathcal{T}(i,j)$ between $i$ and $j$ in $G$. This follows by using
the matrix expansions for paths in $\Lambda^{-1}$ and $K^{-1}$ as we
have used in Sections \ref{sub:pf-directed} and \ref{sub:pf-undirected}.

Similarly, the expansion for $\widetilde{\sigma}_{ij}$ is the sum over all
treks in $\widetilde{G}$. Now set
\[
\phi_{ij} = \widetilde{\phi}_{v_{i,j}, v_{i,j}} \widetilde{\lambda}_{v_{i,j},
i} \widetilde{\lambda}_{v_{i,j}, j} \quad\mbox{and}\quad
\phi_{ii} = \widetilde{\phi}_{ii} + \sum_{j \leftrightarrow i}
\widetilde
{\phi}_{v_{i,j}, v_{i,j}} \widetilde{\lambda}_{v_{i,j}, i}^2.
\]
This transformation shows that these two parametrizations have the same
Zariski closure, since they yield the same formula via sums over the
treks in $G$ and $\widetilde{G}$, respectively. The point is that since we
assume that we are close to the identity matrix, it is also possible to
go back and forth between $G$ and $\widetilde{G}$ parameters. In
particular,\vspace*{1pt} since we are close to the identity matrix, $\phi_{ij}$ is
small. So we can choose $\widetilde{\phi}_{v_{i,j}, v_{i,j}} = \varepsilon>0$
and set $\widetilde{\lambda}_{v_{i,j}, i} = \sqrt{ | \phi_{ij} |
\varepsilon}$ and $\widetilde{\lambda}_{v_{i,j}, j} = \operatorname{sign}( \phi_{ij})
\sqrt{|\phi_{ij} | \varepsilon} $. The small size of the $\phi_{ij}$ guarantee
that we can find a positive $\phi_{ii}$ satisfying the second equation.
The smallness of $\varepsilon$ guarantees that $\Phi$ is positive definite.

(2) Any $t$-separating set in $G$ is clearly a $t$-separating set in
$\widetilde
{G}$. Suppose that $(D_L, D_M, D_R)$ is a minimal $t$-separating set in
$\widetilde{G}$; that is, if any vertex is deleted from $(D_L, D_M, D_R)$
we no longer have a $t$-separating set. It is easy to see that $D_M$ will
not contain any vertices $v_{i,j}$ in a minimal $t$-separating set of
$\widetilde{G}$, so that $D_M\subset V(G)$. It clearly suffices to show
that each minimal $t$-separating set in $\widetilde{G}$ is a $t$-separating set
in $G$. We define
\begin{eqnarray*}
C_L &=& \bigl(D_L \cap V(G)\bigr) \cup\{i \dvtx v_{i,j} \in D_L \}, \\
C_M &=& D_M, \\
C_R &=& \bigl(D_R \cap V(G)\bigr) \cup\{j \dvtx v_{i,j} \in D_R \}.
\end{eqnarray*}
If our $t$-separating set in $\widetilde{G}$ contains none of the vertices
$v_{i,j}$, then it is clearly a \mbox{$t$-separating} set in $G$; otherwise,
the way that $i$ and $j$ are chosen in $\{i \dvtx v_{i,j} \in D_L \}$ and
$\{j \dvtx v_{i,j} \in D_R \}$ is important. Given a vertex $v_{i,j}$ in
the $t$-separating set, let $\mathcal{T}(v_{i,j})$ denote the set of treks
$T=(T_L,T_M,T_R)$ from $A$ to $B$ such that $T_L\cap D_L=\{v_{i,j}\}$
or $T_R\cap D_R=\{v_{i,j}\}$. Since $(D_L, D_M, D_R)$ is minimal, we
see that $\mathcal{T}(v_{i,j})$ must be nonempty. This implies that in every
trek $T=(T_L,T_M,T_R) \in\mathcal{T}(v_{i,j})$, up to relabeling, $i$ occurs
in $T_L$, whose sink lies in $A$, and $j$ occurs in $T_R$, whose sink
lies in $B$. For if there were a trek from $A$ to $B$ in $\mathcal{T}(v_{i,j})$
that had $j$ in $T_L$ or $i$ in $T_R$, we could patch two halves of
these treks together to find a trek from $A$ to $B$ that did not have a
sided intersection with $(D_L, D_M, D_R)$. If $i$ lies in $T_L$ and $j$
lies in $T_R$ in such treks, then we add $i$ to $C_L$ when $v_{i,j} \in
D_L$, and we add $j$ to $C_R$ when $v_{i,j} \in D_R$. Then the triple
$(C_L, C_M, C_R)$ has $\# C_L + \# C_M + \#C_R \leq
\# D_L + \# D_M + \#D_R$ and also $t$-separates $A$ from $B$.
\end{pf}
\begin{rmk*}
The parameterization using the bidirected subdivision $\widetilde{G}$
typically yields a smaller set of covariance matrices than the original
graph $G$. However, these sets have the same dimension and the same
Zariski closure.
\end{rmk*}

Before getting to the general case of mixed graphs, we first need to
handle the special case of mixed graphs that do not have undirected edges.
\begin{lemma}\label{thm:noud}
Suppose that $G$ is a mixed graph without undirected edges. The matrix
$\Sigma_{A,B}$ has rank at most $r$ for all covariance matrices
consistent with the mixed graph $G$ if and only if there exist subsets
$C_L,C_R\subset V(G)$ with $\#C_L + \#C_R \leq r$ such that $(C_L,
\varnothing, C_R)$ $t$-separates $A$ from $B$.
\end{lemma}
\begin{pf}
Due to Proposition \ref{prop:removebi}, this immediately reduces to the
case of directed acyclic graphs, so that we may apply Theorem \ref{thm:main}.
\end{pf}

Now that we have removed the bidirected edges, we assume that our
matrix factorization has the following form:
\[
\Sigma= \Lambda^{-\top}
K^{-1}
\Lambda^{-1}
\]
and we prepare to apply the Cauchy--Binet determinant expansion
formula. That is, for two subsets $A,B \subseteq[m]$, with $\#A = \#
B$, we have
%
%e4 ###
%
\begin{equation} \label{eqn:mixedexpand}
\det\Sigma_{A,B} = \sum_{S \subseteq[m] } \sum_{T \subseteq[n]}
\det
(\Lambda^{-\top})_{A,S} \cdot\det(K^{-1})_{S,T} \cdot\det(
\Lambda
^{-1})_{T,B},
\end{equation}
where the sums range over the sets $S,T \subset[m]$ with $\#S = \#T =
\#
A = \#B$.

We say that a set of treks $\{(P_{L_i}, P_{M_i}, P_{R_i})\dvtx i \in
[\ell]
\}$ has a \textit{sided-crossing} if there are indices $i_1 \neq i_2 \in
[\ell]$ such that either $P_{L_{i_1}}$ and $P_{L_{i_2}}$ share a vertex,
$P_{M_{i_1}}$ and $P_{M_{i_2}}$ share a vertex or
$P_{R_{i_1}}$ and $P_{R_{i_2}}$ share a vertex.
\begin{lemma}\label{lem:mixedcross}
Let $\#A = \# B = r$.
Suppose that every system of $r$ treks from $A$ to $B$ in a mixed graph
$G$ (consisting of directed and undirected edges) has a sided crossing.
Then for every $S, T \subset V(G)$ with $\#S = \#T = r$, we have
$\det(\Lambda^{-\top})_{A,S} \cdot\det(K^{-1})_{S,T} \cdot\det(
\Lambda^{-1})_{T,B} = 0$.
\end{lemma}
\begin{pf}
Consider the trek systems from $A$ to $B$ that consist of a directed
path system $\mathbf{P}_L$ from $S$ to $A$, an undirected path system
$\mathbf{P}_M$ from $S$ to $T$ and a directed path system $\mathbf{P}_R
$ from $T$ to $B$. We call such a system of treks an $(S,T)$-trek
system from $A$ to $B$.

We claim that if every trek system from $A$ to $B$ has a sided
crossing, then either all $(S,T)$-trek systems have a crossing in
$\mathbf{P}_L$, all $(S,T)$-trek systems have a crossing in $\mathbf
{P}_M$ or all $(S,T)$-trek systems have a crossing in $\mathbf{P}_R$.
Suppose this is not the case; then there is a directed path system from
$S$ to $A$ with no crossing, an undirected path system from $S$ to $T$
with no crossing and a directed path system from $T$ to $B$ with no
crossing, yielding an $(S,T)$-trek system from $A$ to $B$ with no sided
crossing.

Applying the claim, along with the directed and undirected versions of
the Gessel--Viennot--Lindstr\"{o}m lemma (Lemma \ref{lem:gvl} and
Corollary \ref{cor:ghat-gvl}), we deduce that one of $\det(\Lambda
^{-\top})_{A,S}$, $\det(K^{-1})_{S,T}$, or $ \det( \Lambda
^{-1})_{T,B}$ is identically zero. This implies that their product is zero.
\end{pf}

Lemma \ref{lem:mixedcross} is enough to handle one direction of Theorem
\ref{thm:mixed}. For the other direction, we need slightly more
machinery. Using our presentation for undirected graphs, we can write
\[
K^{-1} = D^{-1}(I - W)^{-1}D^{-1},
\]
where $D$ is the diagonal matrix of standard deviations, and $W_{ij} =
w_{ij} = w_{ji}$ if $i -j \in E(G)$, and $W_{ij} = 0$ otherwise. Thus,
\[
\Sigma= \Lambda^{-\top} D^{-1} (I - W)^{-1} D^{-1} \Lambda^{-1}.
\]
Using the standard argument of algebraic graph theory, we can expand
this near the identity matrix as a power series,
\[
\sigma_{ij} = \sum_{(P_L, P_M, P_R) \in\mathcal{T}(i,j) } \lambda
^{P_L} d^{-1}_{s(P_L)} w^{P_M} d^{-1}_{s(P_R)} \lambda^{P_R},
\]
where $s(P)$ denotes the source of the directed path $P$.
Thus if $A = \{a_1, \ldots, a_\ell\}$ and $B = \{b_1, \ldots, b_\ell
\},$
%
%e5 ###
%
\begin{eqnarray}\label{eq:biggerexpand}
\det\Sigma_{A,B} &=& \sum_{\tau\in S_\ell, (P_{L_i}, P_{M_i}, P_{R_i})
\in\mathcal{T}(a_{i}, b_{\tau(i)})}
\operatorname{sign}(\tau)\nonumber\\[-8pt]\\[-8pt]
&&{}\times \prod
_{i=1}^\ell
\lambda^{P_{L_i}} d^{-1}_{s(P_{L_i})} w^{P_{M_i}} d^{-1}_{s(P_{R_i})}
\lambda^{P_{R_i}}.\nonumber
\end{eqnarray}
\begin{lemma}\label{lem:mixedcross2}
Suppose that there exists a system of treks from $A=\{a_1,\ldots
,$ $a_\ell \}$
to $B=\{b_1,\ldots,b_\ell\}$ without sided crossing. Then $\det
\Sigma_{A,B}$ is not zero.
\end{lemma}
\begin{pf}
If such a system of treks exists, then there also exists a $\tau\in
S_\ell$ and a system of \textit{simple} treks $T_i=(P_{L_i},P_{M_i},P_{R_i})
\in{\mathcal S}(a_i,b_{\tau(i)}), i=1,\ldots,\ell$ without sided
intersection. Let $G'$ be the graph obtained from $G$ by deleting all
edges that do not appear in any of the $T_i$. The determinant of the
matrix obtained from $\Sigma_{A,B}$ by setting all parameters
corresponding to edges outside
$G'$ equal to zero is exactly the determinant of the corresponding
matrix $\Sigma'_{A,B}$ for $G'$; it suffices to show that this latter
determinant is nonzero.

To do this, we construct a third graph $G''$
from $G'$ by introducing, for each $i$ for which $P_{M_i}$ is not empty,
a bidirected edge $s(P_{L_i}) \leftrightarrow s(P_{R_i})$ with label
$\phi_{s(P_{L_i}),s(P_{M_i})}$ and deleting all undirected edges.
By Lemma \ref{thm:noud} we have $\det\Sigma''_{A,B} \neq0$. But then this
determinant remains nonzero after specialising the parameters
$\phi_{s(P_{L_i}),s(P_{M_i})}$ to the monomials $d_{s(P_{L_i})}^{-1}
w^{P_{M_i}} d_{s(P_{R_i})}^{-1}$; here we use that, as the $P_{M_i}$ are
disjoint, these $\ell$ monomials contain disjoint sets of variables. The
resulting nonzero expression is the subsum of the $G'$-analogue of
(5) over all terms for which the $W$-part of the monomial equals
$\prod_{i=1}^\ell(w^{P_{M_i}})^{\varepsilon_i}$ for some exponents
$\varepsilon_1,\ldots,\varepsilon_\ell\in\{0,1\}$. Indeed, if a system of
treks $(T_i'=(P_{L_i}',P_{M_i}',P_{R_i}'))_i$ from $A$ to $B$ in
$G'$ has $\prod_{i=1}^\ell(w^{P_{M_i}})^{\varepsilon_i}$ as the $W$-part
of its monomial, then since the $P_{M_i}$ are self-avoiding and mutually
disjoint, the nonempty middle parts $P_{M_i}'$ form the subset of the
nonempty $P_{M_i}$ for which $\varepsilon_i$ equals $1$ (potentially up
to traversing
some of these paths in the opposite direction). Hence the trek monomial
of $(T_1',\ldots,T_\ell')$ comes, under the specialization above, from
the monomial of a unique trek in $G''$ of the same sign. This proves
that $\det\Sigma_{A,B}'$ is nonzero, whence the lemma follows.
%\rightqed
\end{pf}
\begin{pf*}{Proof of Theorem \protect\ref{thm:mixed}}
By Proposition \ref{prop:removebi} we can assume that there are no
bidirected edges in $G$.
It suffices to handle the case where $\#A = \#B = r+1$. Lemmas \ref
{lem:mixedcross} and \ref{lem:mixedcross2} imply that $\det\Sigma
_{A,B} = 0$ if and only if every system of $\ell$ treks from $A$ to $B$
has a sided intersection. We wish to apply Menger's theorem. To do
this, we introduce a new graph $\widetilde{G}$ with $3m$ vertices,
namely $\{1,\ldots, m\} \cup\{1', \ldots, m'\} \cup\{1'', \ldots,
m''\}$. This is analogous to our previous definitions of $\widetilde
{G}$, but accounts for both directed and undirected edges. The edge set
of $\widetilde{G}$ consists of precisely those edges,
\begin{itemize}
\item$i\to j$ and $j' \to i'$, where $i \to j$ is a directed edge of $G$,
\item$i'' \to j ''$ and $j''\to i''$, where $i -j$ is an undirected
edge of $G$ and
\item$i'\to i''$ and $i''\to i$, where $i\in[m]$ is a vertex of $G$.
\end{itemize}
Treks between $i$ and $j$ in $G$ are in bijective correspondence with
directed paths between $i'$ and $j$ in $\widetilde{G}$. Thus, the
vertex version of Menger's theorem implies that there must exist $C'_L
\subseteq\{1', \ldots, m'\}$, $C''_M \subseteq\{1'', \ldots, m''\}$
and $C_R \subseteq\{1, \ldots, m\}$ such that every path from $A'$ to
$B$ in $G''$ intersects one of these sets, and such that $\#C'_L + \#
C''_M + \#C_R \leq r$. But then the triple $(C_L, C_M, C_R)$
$t$-separates $A$ from $B$ in $G$ where $C_L=\{c\dvtx c'\in C_L'\}$ and
$C_M=\{c\dvtx c''\in C_M''\}$.
\end{pf*}

%%%%%%%%%%%%%%%%%%%%%%%%%%%%%%%%%
%%%%%%%%%%%%%%%%%%%%%%%%%%%%%%%%%
%%%%%%%%%%%%%%%%%%%%%%%%%%%%%%%%%
%%%%%%%%%%%%%%%%%%%%%%%%%%%%%%%%%

%s4 ###
\section{Conclusions and open problems}\label{sec:questions}

We have shown that the $t$-separation criterion can be used to
characterize vanishing determinants of the covariance matrix in
Gaussian directed and undirected graphical models and mixed graph
models. These results have potential uses in inferential procedures
with Gaussian graphical models, generalizing procedures based on the
tetrad constraints \cite{Scheines1998} in directed graphical models.
The tetrad constraints are the special case of $2 \times2$
determinants. Both referees have pointed out that these results also
extend to graphical models with cycles, by applications of the more
general version of the Gessel--Viennot--Lindstr\"{o}m lemma for general
graphs \cite{Fomin2001}. We have focused on the case of directed
acyclic graphs because these are the most familiar in the graphical
models literature.

Our results suggest a number of different research directions. For
example, for which mixed graphs is it true that vanishing low rank
submatrices characterize the distributions that belong to the model?
This is known to hold for both acyclic directed graphs and undirected
graphs, but can fail in general mixed graphs.

Another open problem is to determine what significance the $t$-separation
criterion has for graphical models with not necessarily normal random
variables, in particular, for discrete variables. It would be
worthwhile to determine whether $t$-separation can be translated into
constraints on probability densities for graphical models with more
general random variables.

\section*{Acknowledgments}

We thank Mathias Drton for suggesting this problem to us. The referees
and associate editor provided many useful comments which have led to
this improved version. Jan Draisma, who was originally an anonymous
referee on this paper, provided the first proof of Theorem 2.17 which
was a conjecture in an earlier version of the paper.

\printaddresses

\end{document}